\documentclass{article}

     \PassOptionsToPackage{numbers, compress}{natbib}

     \usepackage[preprint]{neurips_2023}

\usepackage[utf8]{inputenc} 
\usepackage[T1]{fontenc}    
\usepackage{hyperref}       
\usepackage{url}            
\usepackage{booktabs}       
\usepackage{amsfonts, amsmath, amssymb}       
\usepackage{nicefrac}       
\usepackage{microtype}      
\usepackage{xcolor}         
\usepackage{graphicx}
\usepackage{subcaption}
\usepackage{multirow}
\usepackage{pifont}
\usepackage{float}
\usepackage{amsmath}
\usepackage{array}
\usepackage{wrapfig}
\newcolumntype{P}[1]{>{\centering\arraybackslash}p{#1}}
\newcolumntype{M}[1]{>{\centering\arraybackslash}m{#1}}
\usepackage{color, colortbl}
\definecolor{Gray}{gray}{0.9}

\title{Focus on What Matters: Learning Pose-Aware Video Representations from Vision Transformers}
\title{Focus on the Pose: Learning Pose-Aware Representations in Video Transformers}
\title{Seeing the Pose in the Pixels: Learning Pose-Aware Representations in Vision Transformers}

\author{
  Dominick Reilly\\ 
  UNC Charlotte \\
  \And
  Aman Chadha\thanks{This work is unrelated to the position at Amazon.} \\
  Stanford University, Amazon Alexa AI\\
  \\
   \texttt{dreilly1@charlotte.edu} \\
  \And
 Srijan Das\\
  UNC Charlotte\\
}

\usepackage[draft,textsize=footnotesize,textwidth=15mm]{todonotes}

\newcommand\acb[1]{\textcolor{blue}{#1}}

\newcommand{\red}[1]{\textcolor{red}{#1}}

\begin{document}

\maketitle


\begin{abstract}
    Human perception of surroundings is often guided by the various poses present within the environment. Many computer vision tasks, such as human action recognition and robot imitation learning, rely on pose-based entities like human skeletons or robotic arms. However, conventional Vision Transformer (ViT) models uniformly process all patches, neglecting valuable pose priors in input videos. We argue that incorporating poses into RGB data is advantageous for learning fine-grained and viewpoint-agnostic representations. Consequently, we introduce two strategies for learning pose-aware representations in ViTs.
    The first method, called \textbf{Pose-aware Attention Block (PAAB)}, is a plug-and-play ViT block that performs localized attention on pose regions within videos. The second method, dubbed \textbf{Pose-Aware Auxiliary Task (PAAT)}, presents an auxiliary pose prediction task optimized jointly with the primary ViT task. 
    Although their functionalities differ, both methods succeed in learning pose-aware representations, enhancing performance in multiple diverse downstream tasks. 
    Our experiments, conducted across seven datasets, reveal the efficacy of both pose-aware methods on three video analysis tasks, with PAAT holding a slight edge over PAAB. Both PAAT and PAAB surpass their respective backbone Transformers by up to 9.8\% in \textit{real-world action recognition} and 21.8\% in \textit{multi-view robotic video alignment}. Code is available at \url{https://github.com/dominickrei/PoseAwareVT}.

\end{abstract}

\vspace{-0.2in}

\section{Introduction}
Despite the recent advancements in AI, video understanding remains a formidable task within the field of computer vision.
Transformers~\cite{attention, dosovitskiy2020vit, deit, crossvit, t2t, tnt}, referenced in many recent studies, have proven their dominance across various domains, including vision, thanks to their effective use of self-attention and feed-forward layers. When these transformers are applied to spatio-temporal domains, video transformers~\cite{timesformer, vivit, liu2021videoswin, motionformerNeurIPS21, mvit1} have demonstrated significant potential. However, one limitation is that these video transformers treat all input patches uniformly without accounting for any priors, when applying operations.
This holistic approach is effective for web-sourced videos~\cite{kinetics, caba2015activitynet, ucf,  kuehne2011hmdb}, where prominent motion patterns are typically centered within the image frames. However, the effectiveness of these transformers tends to fall short when employed on daily living videos~\cite{ntu120, NTU_RGB+D, smarthome, MSRDailyactivity3D, nucla} containing subtle motion, non choreographed scenes and varying camera viewpoints. Understanding these videos require learning fine-grained and camera viewpoint agnostic representations.




Daily living videos often contain entities defined by their poses. However, traditional ViTs tend to overlook these pose-based entities during video processing. The effectiveness of 3D pose information is well established in video analysis~\cite{msaagcn, msg3d, skel_CNN_1, skel_CNN_2, skel_cnn_3}. Nevertheless, these pose-based methods are somewhat limited in their ability to model the appearance of a scene.
To address this shortcoming, some studies~\cite{glimpse, STA_hands, das2020vpn} have attempted to combine 3D poses with RGB. Yet, we posit that acquiring 3D poses can be challenging, especially in the absence of a depth sensor, due to the relative inaccuracy of available algorithms and their high computational costs.
\begin{figure}[H]
    \centering
    \scalebox{0.8}{
    \includegraphics[width=\textwidth]{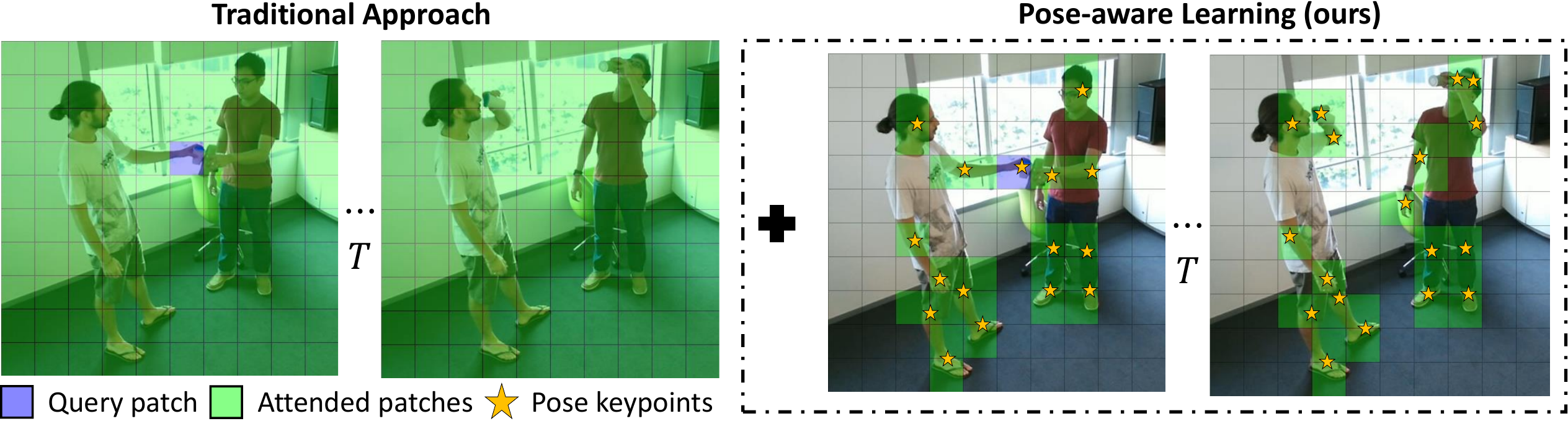}}
    \caption{\textbf{The goal of our proposed framework.} Our aim is to learn pose-aware features for ViTs, while maintaining the whole-scene knowledge learned in the traditional approach to training ViTs.}
    \label{fig:intro} \vspace{-0.15in}
\end{figure}
Consequently, in this paper, we propose learning pose-aware video representations within ViTs by utilizing the capabilities of 2D pose keypoints (see Figure~\ref{fig:intro}). These keypoints, known for their precision, can be easily extracted using readily available pose estimation algorithms~\cite{OpenPose}. To enable the learning of pose-aware representation, we propose two approaches: the first involves the formulation of a network architecture explicitly tailored for localized attention on pose patches and the second introduces an auxiliary pose prediction task that is jointly optimized with the primary task. These approaches has culminated in the introduction of two novel methods: \textbf{Pose-aware Attention Block (PAAB)} and \textbf{Pose-aware Auxiliary Task (PAAT)}. Both PAAB and PAAT can be plugged into an existing ViT, with an auxiliary loss being added for the latter. Despite their differing functionalities, both PAAB and PAAT learn to disentangle between pose and non-pose patches within a video. Our analysis leads us to the striking observation that the learned pose-aware representations are not merely a result of the pose-guided sparsity of the ViT's attention weights, but are instead achieved through the feed-forward layers within the ViT. The efficacy of PAAB and PAAT is validated across three downstream tasks encompassing three datasets for action recognition, four for multi-view robotic video alignment, and a cross-data evaluation for video retrieval. Both PAAB and PAAT significantly outperform the baseline Transformer across all datasets, achieving state-of-the-art results relative to representative baselines.


\section{Background: Attention in Video Transformers}
Pose-aware representation learning is based on incorporating pose information into the training process of existing ViTs. As such, we briefly review how attention is performed in ViT's for video data.
Consider a video input of size $\tau \times H \times W \times 3$, where $\tau$ frames have a spatial resolution of $ H \times W$ and three color channels. Most video transformers \cite{Selva2022video_transformer_survey} extract disjoint patches from the video, resulting in an input sequence of $ST$ tokens, with $S$ being the spatial resolution and $T$ being the temporal resolution. Each of these tokens are then projected to $\mathbb{R}^D$ via a linear layer. Subsequently, two learnable position embeddings 
are added to each token to encode spatial and temporal position information, respectively. Furthermore, a class token is added to the input sequence prior to its processing by the transformer to enable the classification of the entire video. Note that this class token can be used for performing other downstream tasks as well.

Similarly to the standard transformer, the input sequence is transformed into key, query, and value matrices denoted as $\mathbf{K} \in \mathbb{R}^{ST \times D}$, $\mathbf{Q} \in \mathbb{R}^{ST \times D}$, and $\mathbf{V} \in \mathbb{R}^{ST \times D}$, respectively. Conventional self-attention \cite{attention} computes the pairwise similarities between all combinations of tokens in the input sequence. In the realm of video transformers this is known as \textit{joint space-time attention}, as similarity is computed between all tokens, regardless of their spatial or temporal position as: 
\vspace{-0.025in}
\begin{equation} 
    \boldsymbol{\alpha}_{st}^{\mathrm{joint}} = \frac{\mathrm{exp}(\mathbf{Q}_{st} \mathbf{K}^{\top})}{\sum_{s't'}\mathrm{exp}(\mathbf{Q}_{st} \mathbf{K}_{s't'}^{\top})} \\
    \label{joint_normal_attention}
\end{equation}
where, $\mathbf{Q}_{st}$, $\mathbf{K}_{s't'}$ are the $D$-dimensional query and key vector for the token at spatial and temporal position $s, t$ respectively. 
However, this approach for computing attention in video transformers is expensive due to the quadratic complexity of self-attention and the large size of video data. To address this, factorized self-attention has been proposed in~\cite{timesformer}. 
This mechanism is termed \textit{divided space-time attention} and is achieved by applying temporal-attention followed by spatial-attention as: 
\vspace{-0.05in}
\begin{equation}
    \boldsymbol{\alpha}_{st}^{\mathrm{time}} = \frac{\mathrm{exp}(\mathbf{Q}_{st} \mathbf{K}_{s:}^\top)}{\sum_{t'}\mathrm{exp}(\mathbf{Q}_{st} \mathbf{K}_{st'}^\top)};
    \hspace{0.2in}
    \boldsymbol{\alpha}_{st}^{\mathrm{spatial}} = \frac{\mathrm{exp}(\mathbf{Q}_{st} \mathbf{K}_{:t}^\top)}{\sum_{s'}\mathrm{exp}(\mathbf{Q}_{st} \mathbf{K}_{s't}^\top)}
    \label{spatial_temporal_normal_attention}
\end{equation}

where, $\mathbf{K}_{:t}$ indicates a slice of $\mathbf{K}$ across the $t^{th}$ frame (i.e., the keys for all spatial tokens in frame $t$). The remaining operations within video transformers follow the same principles as standard vision transformers~\cite{dosovitskiy2020vit}.

\section{Pose-Aware Representation Learning}
This section presents our approaches to learning pose-aware representations by utilizing existing Vision Transformer (ViT) architectures for video understanding. Towards this objective, we propose two distinct approaches. The first approach entails introducing architectural changes through the incorporation of a novel PAAB that integrates knowledge of poses into the ViT representation. In contrast, the second approach involves the use of PAAT, a multi-tasking objective function to reinforce the ViT's focus on poses, facilitating the learning of pose-aware representations.

\subsection{Pose map instantiations}\label{pose_map_init}
Due to the nature of our methods, we must have a correspondence between the video patches and the pose configurations of objects within the video. We achieve this through the construction of two pose maps: $\mathcal{P}^{2D}$ and $\mathcal{P}^{3D}$, the details of which are described below.

The pose configuration of an object in a video is typically characterized by its 2D pose, which is represented by a set of 2D coordinates (known as keypoints) that provides the specific locations of relevant parts of the object. For instance, in human action videos these keypoints correspond to the locations of various human joints (hand, foot, etc) in each video frame. The localization of these keypoints can be achieved by exploiting pose estimation algorithms such as \cite{OpenPose, fang2017rmpe}, which are high precision algorithms and are commonly used in video analysis. 
After extracting these keypoints, we obtain a set $\mathcal{K}$ that denotes the coordinates of the keypoints within each frame:
\begin{equation}
    \mathcal{K} = \{(t, k, x, y)\} : 1 \leq t \leq \tau, 1 \leq k \leq K
\end{equation}

where $K$ is the number of keypoints. 
We then define a pose map, $\mathcal{P}$, of resolution $\tau \times K \times H \times W$ as:
\vspace{-0.025in}
\begin{equation}
    \mathcal{P}_{tkxy} = 
    \begin{cases}
        1 & \textrm{if} \: (t, k, x, y) \in \mathcal{K}\\
        0 & \textrm{otherwise}
    \end{cases}
    \label{}
\end{equation}

Thus, $\mathcal{P}_{tkxy} = 1$ if the $k^{th}$ keypoint is present at the $(x,y)^{th}$ pixel in the $t^{th}$ video frame. To align with ViT inputs, $\mathcal{P}$ is decomposed into $ST$ disjoint patches, transforming $\mathcal{P}$ into a $K \times ST \times p \times p$ dimensional binary matrix, where $p$ is the patch size. Each patch is transformed as $\mathcal{P}^{2D}_i = \mathrm{MaxPool}(\mathcal{P}_i)$, i.e., if the patch $\mathcal{P}_i$ contains one or more keypoints, it is set to one, else it remains zero. Thus, $\mathcal{P}^{2D}$ is a $ST$ dimensional binary vector indicating the video patches that contain \textit{any} keypoints.


We also compute a 3D instantiation of the pose map, denoted as $\mathcal{P}^{3D}$. In contrast to $\mathcal{P}^{2D}$, $\mathcal{P}^{3D}_{ik} = 1$ if the $k^{th}$ keypoint lies in the patch $\mathcal{P}_i$ , otherwise it is zero.
Thus, $\mathcal{P}^{3D}$ is a $ST \times K$ dimensional binary vector indicating video patches that contain \textit{a specific} keypoint.

\begin{figure*}
    \begin{subfigure}{0.44\textwidth}
        \includegraphics[width=\textwidth]{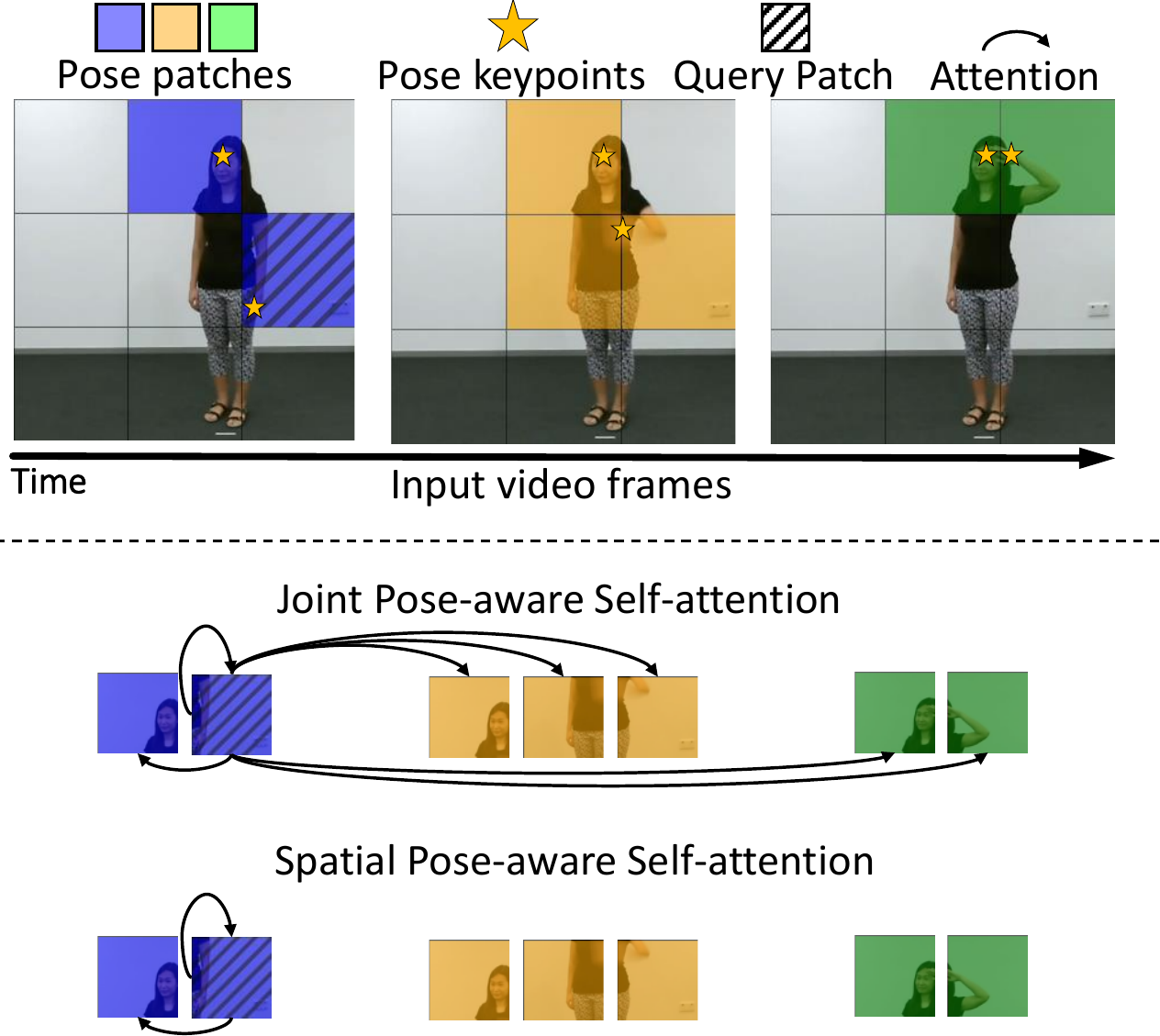}
        \caption{Visual of pose-aware attention schemes.}
        \label{fig:paab_inputs_and_attentions}
    \end{subfigure}
    \hfill
    \begin{subfigure}{0.52\textwidth}
        \includegraphics[width=\textwidth]{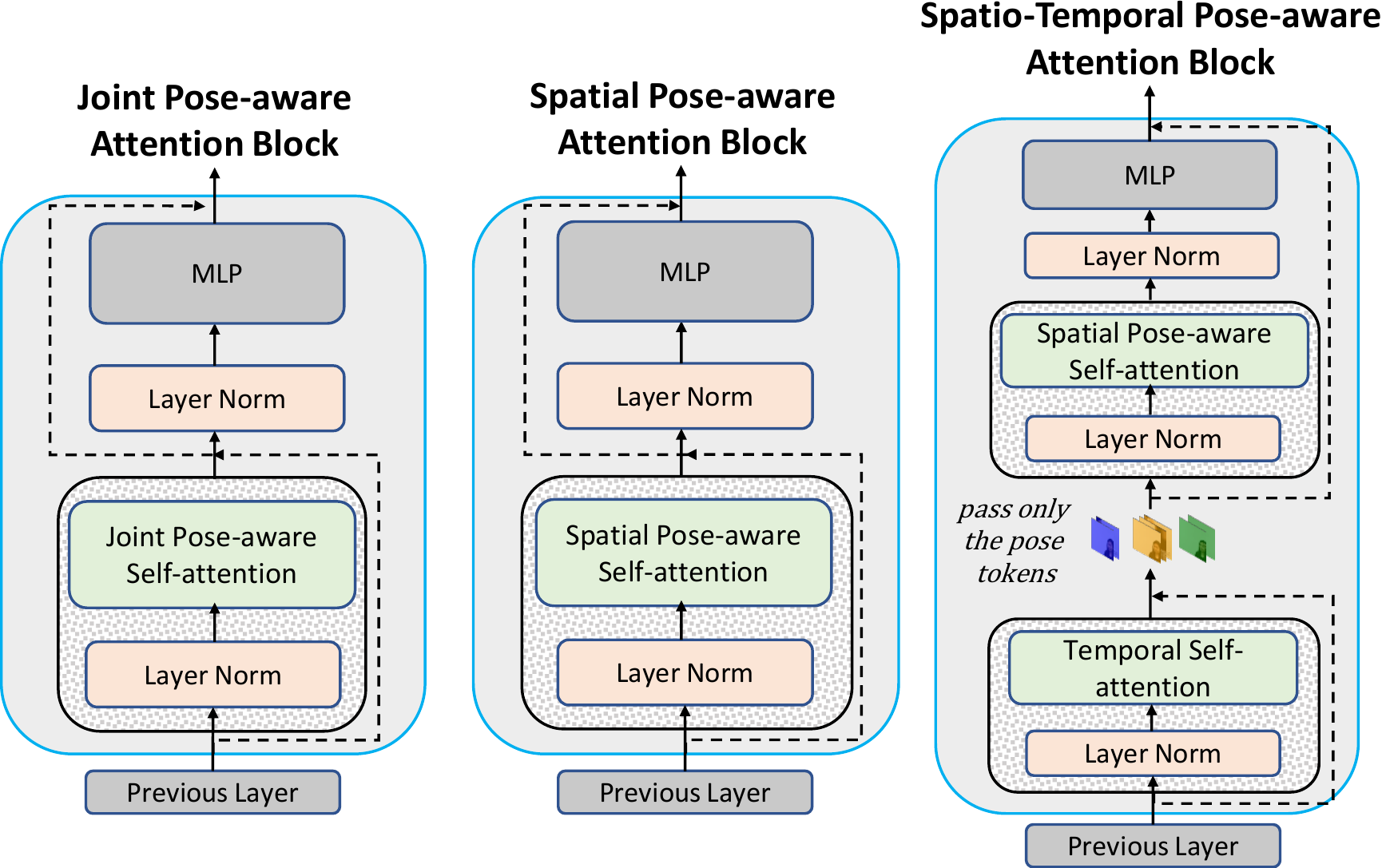}
        \caption{Variants of Pose-aware Transformer Block. Dashed lines indicated residuals.}
        \label{fig:paab_variants}
    \end{subfigure}
    \caption{\textbf{Overview of Pose-Aware Attention Block.} PAAB takes tokens processed by a ViT as input and applies a pose-aware attention to them. The attention is applied jointly (over pose tokens from all frames), spatially (over pose tokens in a single frame), or spatio-temporally. In spatio-temporal attention, traditional temporal attention is applied followed by pose aware spatial attention.} \vspace{-0.25in}
\end{figure*}

\subsection{Pose-Aware Attention Block (PAAB)}\label{sec:paab_introduction}
The Pose-Aware Attention Block (PAAB) is a plug-in module that can be inserted into existing ViT architectures to induce learning of pose-aware representations. When inserted into a ViT, PAAB will process tokens from the previous layer and return a set of tokens enriched with pose information, these enriched tokens can be propagated as usual through the rest of the ViT. PAAB accomplishes this through a pose-aware self-attention mechanism, restricting interactions to tokens representing human keypoints, i.e., \textit{pose tokens}. Essentially, PAAB functions as a local attention that modulates the pose token representation based on its interaction with other pose tokens within a video. As shown in Figure \ref{fig:paab_variants}, PAAB comes in three variants, namely, joint (Joint PA-STA), spatial (PA-SA), and spatio-temporal (Factorized PA-STA) pose-aware self-attention, each differing based on how pose tokens interact with each other.

The joint variant of PAAB (see Figure~\ref{fig:paab_inputs_and_attentions}) extends the joint space-time attention in equation~\ref{joint_normal_attention} by leveraging the 2D pose map $\mathcal{P}^{2D}$  as
\vspace{-0.05in}
\begin{equation} 
    \boldsymbol{\alpha}_{st}^{\mathrm{PA-joint}} = 
    \begin{cases}
        \frac{\mathrm{exp}(\mathbf{Q}_{st} \mathbf{K}^\top \odot \boldsymbol{\mathcal{P}}^{2D})}{\sum_{s't'}\mathrm{exp}(\mathbf{Q}_{st} \mathbf{K}_{s't'}^\top \odot \boldsymbol{\mathcal{P}}_{s't'}^{2D})} & \textrm{if} \: \mathcal{P}^{2D}_{st} = 1 \\
        \mathbf{0} & \textrm{if} \: \mathcal{P}^{2D}_{st} = 0
    \end{cases}
    \label{joint_poseaware_attention}
\end{equation}
where $\odot$ is the Hadamard product. Similarly, a spatial variant of PAAB learns attention weights $\boldsymbol{\alpha}_{st}^{\mathrm{PA-spatial}}$ for the token at ($s,t$) as, 
\vspace{-0.05in}
\begin{equation}
    \boldsymbol{\alpha}_{st}^{\mathrm{PA-spatial}} = 
    \begin{cases}
        \frac{\mathrm{exp}(\mathbf{Q}_{st} \mathbf{K}_{:t}^\top \odot \boldsymbol{\mathcal{P}}_{:t}^{2D})}{\sum_{s'}\mathrm{exp}(\mathbf{Q}_{st} \mathbf{K}_{s't}^\top \odot \boldsymbol{\mathcal{P}}_{s't}^{2D})}  & \textrm{if} \: \mathcal{P}^{2D}_{st} = 1 \\
        \mathbf{0} & \textrm{if} \: \mathcal{P}^{2D}_{st} = 0
    \end{cases}
    \label{spatial_poseaware_attention}
\end{equation}
Using $\mathcal{P}^{2D}$, the spatial attention in PAAB allows interaction amongst pose tokens in a single frame (see Figure~\ref{fig:paab_inputs_and_attentions}). The spatio-temporal variant of PAAB entails temporal attention $ \boldsymbol{\alpha}^{\mathrm{time}}$, applied to all tokens as described by equation~\ref{spatial_temporal_normal_attention}. This is then followed by a pose-aware spatial attention $\boldsymbol{\alpha}^{\mathrm{PA-spatial}}$, as depicted by equation~\ref{spatial_poseaware_attention}.
Note that the aforementioned conditional pose-aware attention weights are implemented in practice through a differential approximation, whereby $\mathcal{P}^{2D} = \infty(\mathcal{P}^{2D}-1)$ is transformed, followed by adding (instead of Hadamard product) with $\mathbf{Q}_{st} \mathbf{K}^\top$, as performed in equation~\ref{joint_poseaware_attention} and~\ref{spatial_poseaware_attention}. This equates to masking out the attention values of all non-pose tokens, similarly to how the decoder in NLP transformers masks out unseen tokens~\cite{attention}.




\subsection{Pose-Aware Auxiliary Task (PAAT)}\label{sec:paat_introduction}
\begin{figure*}
    \begin{subfigure}{0.45\textwidth}
        \includegraphics[width=\textwidth]{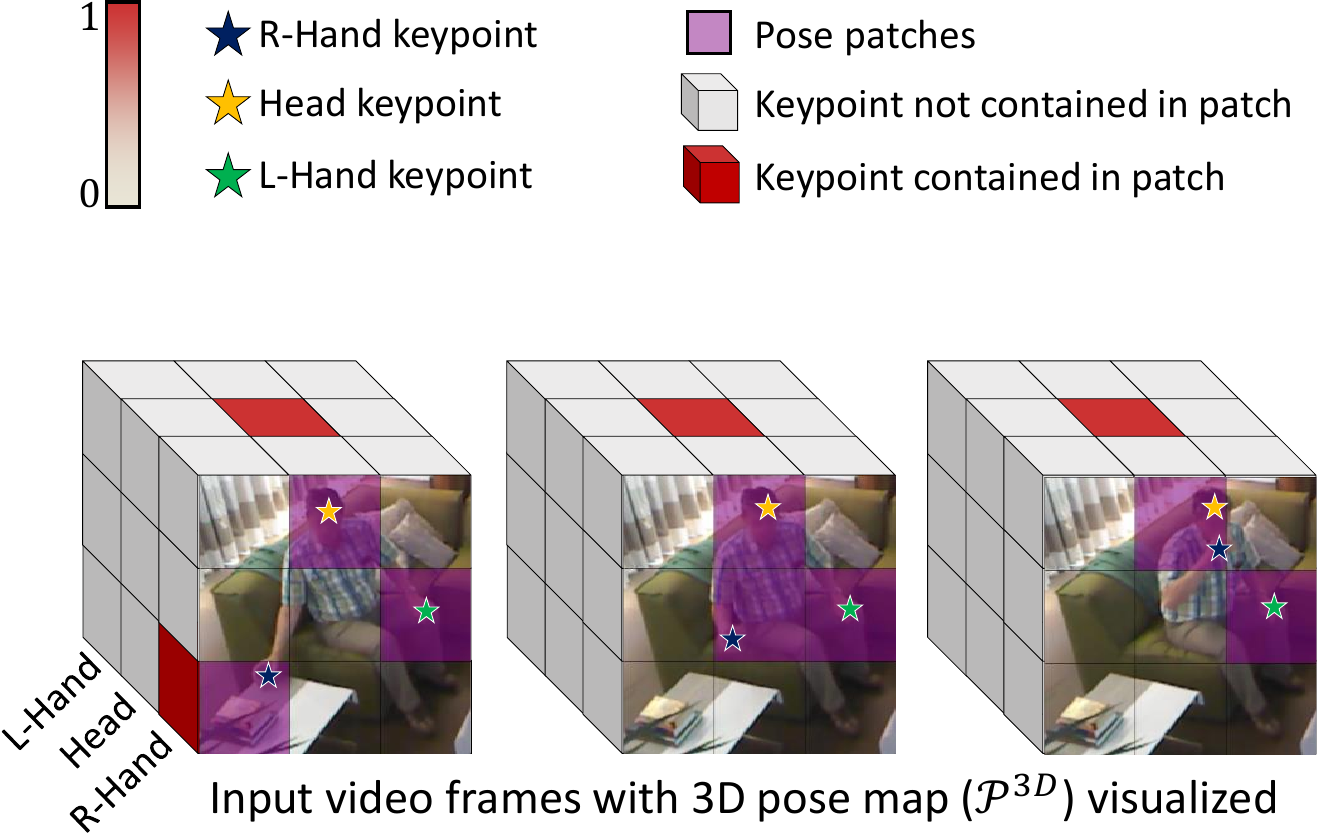}
        \caption{3D Pose maps generated from the video frames}
        \label{fig:paat_inputs_and_attentions}
    \end{subfigure}
    \hfill
    \begin{subfigure}{0.50\textwidth}
        \includegraphics[width=\textwidth]{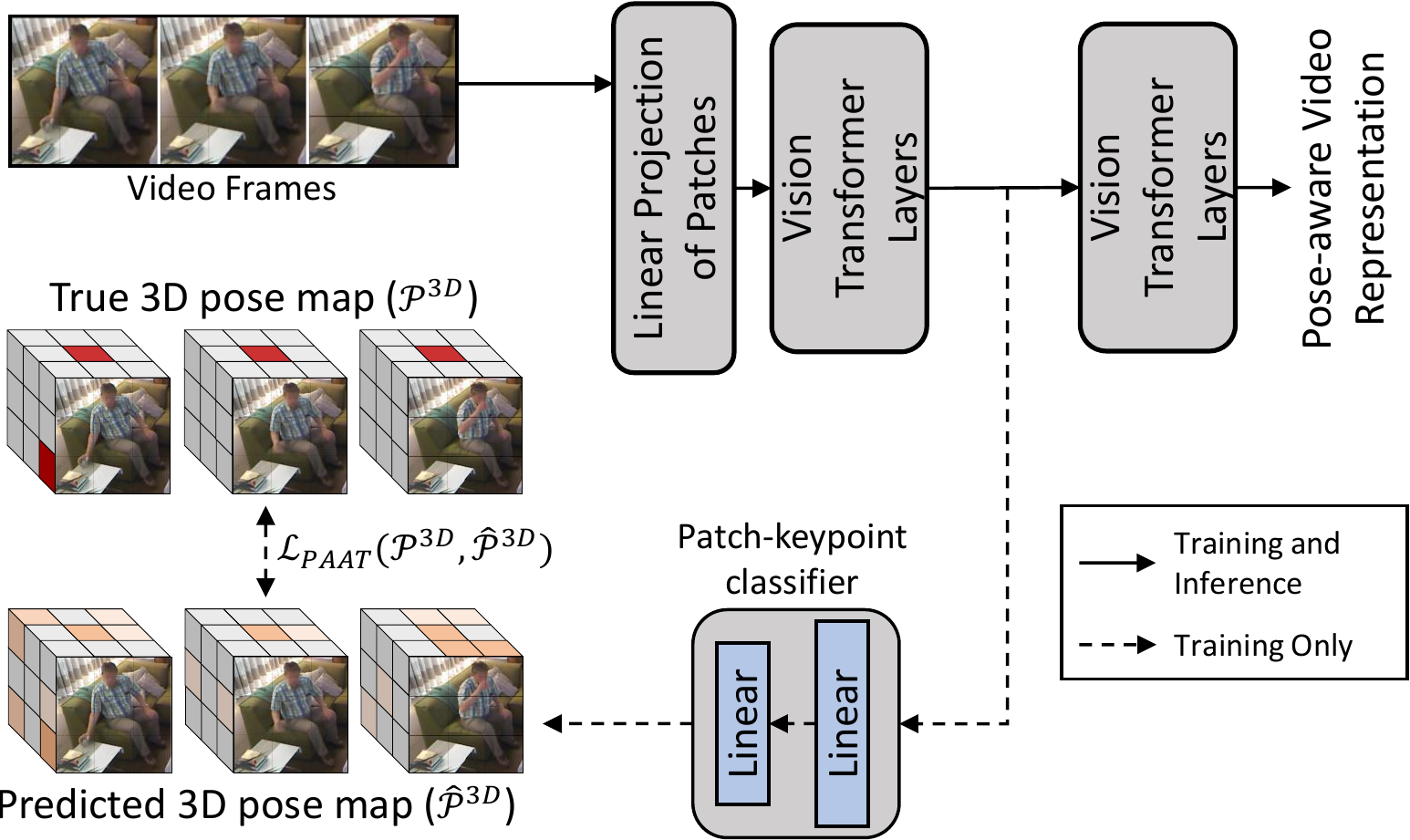}
        \caption{Pipeline}
        \label{fig:paat_variants}
    \end{subfigure}
    \caption{\textbf{Overview of Pose-Aware Auxiliary Task.} Given the 3D pose map, PAAT learns to predict the specific keypoint present within each video patch. This task is learned at train time via the patch-keypoint classifier, and can be discared at inference time.} \vspace{-0.15in}
\end{figure*}

In contrast to PAAB, which utilizes a local attention mechanism to facilitate pose-aware representation learning, PAAT attempts at achieving the same through the introduction of an auxiliary task that is jointly optimized alongside the primary ViT task. 
When inserted into a ViT, PAAT's goal is to classify the specific keypoints present in each patch using the intermediate token representations obtained from the ViT layer preceeding PAAT. In other words, its goal is to predict the 3D pose map $\mathcal{P}^{3D}$. This task can be realized as a multi-label multi-class classification problem, as each patch can contain multiple keypoints (illustrated in Figure~\ref{fig:paat_inputs_and_attentions}).

Given a set of tokens from the layer preceeding PAAT, $\mathbf{z}_{l-1} \in \mathbb{R}^{ST \times D}$, PAAT predicts the 3D pose map introduced in section~\ref{pose_map_init}. Recall that each $D$-dimensional token in $\mathbf{z}_{l-1}$ corresponds to the latent representation of a video patch from the $(l-1)^{th}$ ViT layer.
As depicted in Figure~\ref{fig:paat_variants}, PAAT is formulated as a \textit{patch-keypoint classifier} that is composed of two linear layers defined by the weights $\mathbf{W}_1 \in \mathbb{R}^{D \times D_e}$ and $\mathbf{W}_2 \in \mathbb{R}^{D_e \times K}$, where $D_e$ is the bottleneck dimension and $D_e \leq D$. The 3D pose map predicted by PAAT inserted at $l^{th}$ layer of a ViT is given by:
\vspace{-0.05in}
\begin{equation}
    \hat{\mathcal{P}}^{3D} = \sigma((\mathbf{z}_{l-1} \mathbf{W}_1) \mathbf{W}_2)
\end{equation}
where $\sigma$ is the sigmoid activation. For brevity, we omit the bias terms. 
PAAT's loss is computed as the binary cross-entropy (BCE) between $\mathcal{P}^{3D}$ and $\hat{\mathcal{P}}^{3D}$. During training, PAAT is optimized jointly with the primary task, such as classification or video alignment, with a loss $\mathcal{L}_{primary}$. Consequently, PAAT loss, $\mathcal{L}_\mathrm{PAAT}$, and model's total loss, $\mathcal{L}_{\mathrm{total}}$, is defined as,
\vspace{-0.05in}
\begin{equation}
    \mathcal{L}_\mathrm{PAAT} = \mathrm{BCE}(\mathcal{P}^{3D}, \hat{\mathcal{P}}^{3D}); \quad 
    \mathcal{L}_\mathrm{total} = \lambda\mathcal{L}_{PAAT} + \mathcal{L}_{primary}
\end{equation}
\vspace{-0.05in}
where $\lambda$ is a scaling factor that controls the influence of PAAT on the model training. At training, the gradient updates with $\frac{\partial \mathbf{z}_{l-1}}{\partial \mathcal{L}_\mathrm{PAAT}}$ at the $(l-1)^{th}$ layer force the ViT to learn pose-aware representations that discriminate between pose and non-pose tokens. This enables the remaining transformer layers to encode pose-aware representations. At inference, the patch-keypoint classifier is discarded and the ViT can be used with no remnants of PAAT.

\section{Experiments}
We evaluate the effectiveness of the proposed pose-aware learning methods on three diverse computer vision tasks: (i) action recognition, (ii) multi-view robotic video alignment, and (iii) video retrieval. We provide an experimental analysis that demonstrates the effectiveness of our methods across these diverse tasks. We also perform an extensive diagnosis on our model and discuss the intriguing properties we observe.
 
\subsection{Datasets \& Evaluation protocols}
\textbf{Action recognition} is a popular video analysis task whose goal is to learn to predict an action label given a trimmed video.
For this task we evaluate our methods on three popular Activities of Daily Living (ADL) datasets: \textbf{Toyota-Smarthome} \cite{smarthome} (Smarthome, SH), \textbf{NTU-RGB+D} \cite{NTU_RGB+D} (NTU), and \textbf{Northwestern-UCLA Multiview activity 3D Dataset} \cite{nucla} (NUCLA). For the Toyota-Smarthome dataset, we adhere to the cross-subject (CS) and cross-view (CV1, CV2) protocols, gauging performance using the mean class-accuracy (mCA) metric. When assessing the NTU-RGB+D dataset, we follow the cross-view-subject (CVS) protocols proposed in \cite{varol21_surreact}, as they better represent the cross-view challenge. As for the NUCLA dataset, we report the accuracy on cross-subject (CS), cross-view (CV3), and the average across all the cross-view protocols. For the extraction of 2D pose keypoints, we employed LCRNet~\cite{lcrnet_new}, Randomized Decision Forest~\cite{37}, and OpenPose~\cite{OpenPose} for Smarthome, NTU and NUCLA datasets respectively. Note that all our ablation studies are conducted on the action classification task.

\textbf{Multi-view robotic video alignment} is a task to learn a frame-to-frame mapping between video pairs acquired from different camera viewpoints. Such tasks are able to facilitate robot imitation learning from third-person viewpoints~\cite{3dtrl}.
For this task we use the \textbf{Minecraft} (MC), \textbf{Pick}, \textbf{Can}, and \textbf{Lift} datasets. These datasets are obtained from a range of environments: Minecraft from video game whereas Pick, Can, and Lift from robotics simulators (PyBullet\cite{pybullet}, Robomimic\cite{robomimic2021}).The pixel positions of the robotic arms, regarded as the poses, are obtained from the simulators. This task is evaluated by an alignment error metric introduced in~\cite{sermanet2017tcn_robotalign}. Sample frames from these datasets are provided in Fig.~\ref{fig:alignment_datasets}. 


\textbf{Video Retrieval} is a nearest-neighbour retrieval task performed on learned features without any further training. 
For evaluation, we report the Recall at $k$ ($R@k$), meaning, if the top k nearest neighbours contains one video of the same class, the retrieval was successful.


\subsection{Implementation}
In our default implementations of PAAB, we use the spatial attention variant (PA-SA) inserted after the $12^{th}$ layer of a backbone ViT. For PAAT, we insert it after the $1^{st}$ layer of the backbone. For the implementation of PAAT, we use a bottleneck dimension ($D_e$) of $256$ for the patch-keypoint classifier and a loss scale ($\lambda$) of $1.6$. In our experiments, we use a TimeSformer~\cite{timesformer} backbone for the task of action recognition and video retrieval, while a DeiT~\cite{deit} backbone is utilized for video alignment. 
All other training and dataset specific details are provided in the Appendix. 

\begin{table}
    \centering
    \caption{\textbf{Ablations on PAAB and PAAT.} We perform ablations on the following: (a) position of PAAB and PAAT, (b) variants of PAAB, (c) number of PAABs to insert, and (d) variants of PAAT.}
    \begin{subfigure}{0.9\textwidth}
        \caption{PAAB performs best when inserted near the end of the model, PAAT performs best when inserted at the beginning. Inserting the PAAB or PAAT at multiple positions is not necessary.}
        \resizebox{\textwidth}{!}{
        \begin{tabular}{c|c|ccccc|ccccc}
            \toprule
            \multirow{2}{*}{\textbf{Dataset}} & \multirow{2}{*}{\textbf{Baseline}} & \multicolumn{5}{c|}{\textbf{PAAB Position}} & \multicolumn{5}{c}{\textbf{PAAT Position}} \\
             & & \textbf{1} & \textbf{6} & \textbf{12} & \textbf{1,6} & \textbf{1,12} & \textbf{1} & \textbf{6} & \textbf{12} & \textbf{1,6} & \textbf{1,12} \\
            \toprule
            SH (CS) & 68.4 & 67.1 & 67.7 & \textbf{71.4} & 67.1 & 68.6 & \textbf{72.5} & 70.9 & 69.9 & 70.7 & 70.2 \\
            SH (CV1) & 50.0 & 50.5 & 52.4 & \textbf{54.9} & 51.5 & 50.5 & \textbf{54.8} & 52.2 & 47.6 & 49.7 & 51.5\\
            \midrule
            NTU (CVS1) & 83.5 & 85.0 & \textbf{85.7} & 85.2 & 85.5 & 85.2 & \textbf{85.4} & 85.2 & 84.5 & 84.6 & 84.2\\
            \bottomrule
        \end{tabular}
        }
		\label{tab:PATB_PAAT_position_ablation}
    \end{subfigure}

    \begin{subfigure}{0.35\textwidth}
        \vspace{0.1cm}
        \caption{PA-SA is sufficient despite having the least  additional parameters.}
        \resizebox{\textwidth}{!}{
        \begin{tabular}{ccccc}
            \toprule
             \multirow{2}{*}{\textbf{Dataset}} & \multirow{2}{*}{\textbf{PA-SA}} & \textbf{Factorized} & \textbf{Joint}  \\
              &  & \textbf{PA-STA} & \textbf{PA-STA} \\
            \toprule
            SH (CS) & \textbf{71.4} & 69.9 & 69.8 \\
            SH (CV1) & \textbf{54.9} & 50.2 & 52.0 \\
            \midrule
            NTU (CVS1) & 85.2 & 85.4 & \textbf{85.8} \\
            NTU (CVS3) & \textbf{51.6} & 51.3 & 49.2 \\
            \bottomrule
        \end{tabular}
        }
	  \label{tab:PAAB_variant_ablation}
    \end{subfigure}
    \hfill
    \begin{subfigure}{0.36\textwidth}
        \caption{Inserting $1$ PAAB after layer 12 is the most consistent across datasets.}
        \resizebox{\textwidth}{!}{
    	\begin{tabular}{cc c c c cc c c c}
            \toprule
            \multirow{2}{*}{\textbf{Dataset}} & \multicolumn{4}{c}{\textbf{\# PAABs at layer 12}} \\
              & \textbf{1} & \textbf{2} & \textbf{3} & \textbf{4} \\
            \toprule
            SH (CS) & \textbf{71.4} & 67.1 & 69.0 & 66.7 \\
            SH (CV1) & \textbf{54.9} & 51.3 & 51.18 & 48.4 \\
            \midrule
            NTU (CVS1) & 85.2 & 84.7 & 85.0 & \textbf{85.6}  \\
            \bottomrule
        \end{tabular}
        }
	  \label{tab:num_PAAB}
    \end{subfigure}
    \hfill
    \begin{subfigure}{0.27\textwidth}
        \caption{PAAT performs best when predicting $\mathcal{P}^{3D}$ over $\mathcal{P}^{2D}$}
        \resizebox{\textwidth}{!}{
        \begin{tabular}{cc c c}
            \toprule
             \textbf{Dataset} & \textbf{Variant} & \textbf{Accuracy}  \\
            \toprule
            \multirow{2}{*}{SH (CS)} & $\mathcal{P}^{2D}$ & 71.1 \\
            & $\mathcal{P}^{3D}$ & \textbf{72.5} \\
            \midrule
            \multirow{2}{*}{NTU (CVS1)} & $\mathcal{P}^{2D}$ & 84.0 \\
            & $\mathcal{P}^{3D}$ & \textbf{85.4} \\
            \bottomrule
        \end{tabular}
        }
		\label{tab:PAAT_variant_ablation}
    \end{subfigure}
    
    \label{tab:paab_paat_ablation} \vspace{-0.2in}
\end{table}

\subsection{Ablation studies}
\textbf{Where should we insert PAAB and PAAT?}\quad
In Table \ref{tab:PATB_PAAT_position_ablation}, we investigate the optimal insertion point of PAAB and PAAT. We initially examine the performance impact of incorporating a single PAAB or PAAT after specific ViT layers. Subsequently, we assess the implications of integrating multiple PAABs or PAATs at varying ViT layers. 
Interestingly, our findings suggest a complementary dynamic between PAAB and PAAT. PAAB exhibits superior performance when positioned closer to the ViT's classification head, while PAAT performs better when inserted at the initial layer. 
This shows that the auxiliary task is beneficial for improving the primary task, namely action recognition, when operating on low-level token representations that have not yet been contextualized~\cite{attn_rollout}.
In contrast, attention blocks like PAAB are most effective when working with high-level token representations, which have been extensively contextualized and optimized for the primary task.


\textbf{Which variant of PAAB and PAAT should be used?}\quad
Here, we explore the different variants of PAAB and PAAT. Table~\ref{tab:PAAB_variant_ablation} presents the classification results of different PAAB variants. These attention variants are arranged from left to right, corresponding to the number of extra parameters they introduce. Across the Smarthome and NTU datasets, PA-SA consistently exhibits superior performance, while the other two variants tend to result in a significant decrease in performance (for instance, a reduction of $2.81\%$ on Smarthome (CS) when transitioning from PA-SA to Joint PA-STA). In general, despite having the lowest number of added parameters, PA-SA proves to be sufficient for learning pose-aware representations.
In Table~\ref{tab:PAAT_variant_ablation}, we analyze the implications of training PAAT with an alternate auxiliary task. This task involves predicting the presence or absence of a keypoint in each patch, rather than the specific keypoint located in each patch. Essentially, the task's objective is to predict the 2D pose map instantiation, $\mathcal{P}^{2D}$. We ascertain that the patch-keypoint prediction task (predicting $\mathcal{P}^{3D}$), proves more effective, underlining the significance of incorporating human anatomy knowledge into the learned video representation.

\textbf{How many PAAB's should you use?}\quad
In Table \ref{tab:num_PAAB}, we examine the optimal number of consecutive PAABs to be incorporated into the ViT. 
Our findings suggest that a single PAAB is sufficient, and the model's performance tends to decline with an increased number of blocks. This outcome is attributed to the fact that incorporating additional PAABs leads to a loss of the valuable contextual information from non-pose tokens, which are often crucial for action recognition.

\begin{figure}[htbp]
    \vspace{-3pt}
    \begin{minipage}{0.5\textwidth}
        \centering
        \setlength{\tabcolsep}{1pt}
        \captionof{table}{Results of training our models with and without random 2D and 3D pose maps.}
        \resizebox{\textwidth}{!}{
        \begin{tabular}{c M{2.8cm} M{2.2cm} M{2.8cm}}
            \toprule
            \textbf{Dataset} & \textbf{Method} & \textbf{Random pose map} & \textbf{Accuracy (\%)}  \\
            \toprule
            \multirow{4}{*}{SH (CS)} & \multirow{2}{*}{PAAB} & \ding{51} & 70.1 \\
            & & \ding{55} & 71.4 \\
            \cline{2-4}
            & \multirow{2}{*}{PAAT} & \ding{51} & 69.5 \\
            & & \ding{55} & 72.5 \\
            \midrule
            \multirow{4}{*}{NTU60 (CVS1)} & \multirow{2}{*}{PAAB} & \ding{51} & 85.0 \\
            & & \ding{55} & 85.8 \\
            \cline{2-4}
            & \multirow{2}{*}{PAAT} & \ding{51} & 84.8 \\
            & & \ding{55} & 85.4 \\
            \bottomrule
        \end{tabular}
        }
        \label{tab:random_keypoints}
    \end{minipage}
    \hfill
    \begin{minipage}{0.4\textwidth}
        \includegraphics[width=\textwidth]{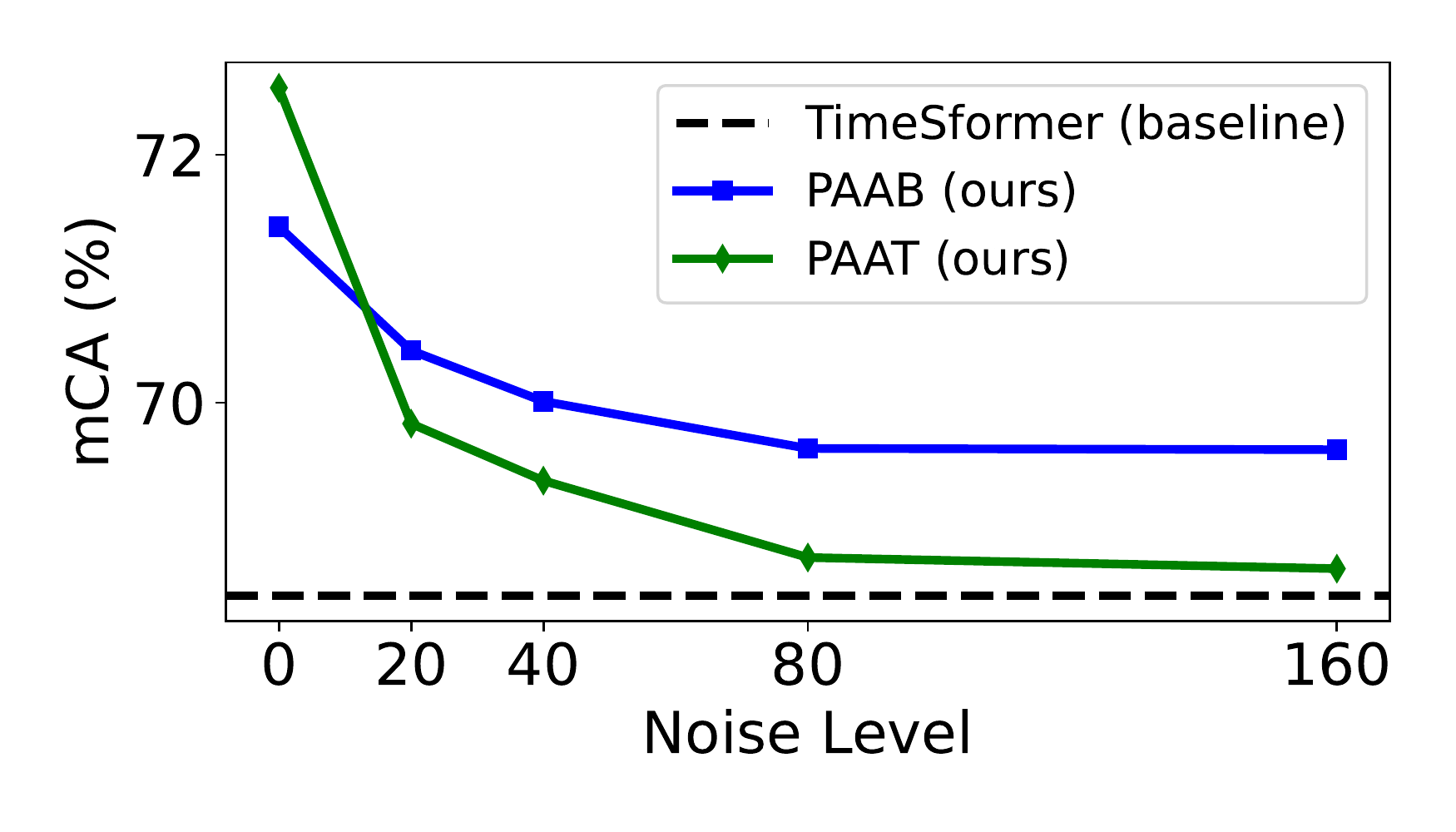}
        \vspace{-0.6cm}
        \captionof{figure}{Model performance degrades as we lose more of the pose information.}
        \label{fig:noisy_poses}
    \end{minipage}
    \vspace{-10pt}
\end{figure}

\subsection{Do poses really help?}

To answer this question, we perform two experiments to evaluate the importance of poses. In our first experiment, we randomly activate values in the pose maps, $\mathcal{P}^{2D}$ and $\mathcal{P}^{3D}$, irrespective of the actual presence or absence of pose keypoints in a patch. Our results, presented in Table \ref{tab:random_keypoints}, show that our methods deliver superior performance when utilizing accurate pose maps informed by pose estimation as opposed to random pose maps.
In the second experiment, we introduce varying levels of noise to the pose keypoints before computing $\mathcal{P}^{2D}$ and $\mathcal{P}^{3D}$. More specifically, we set a noise level $\epsilon \geq 0$ and add a randomly generated integer between $0$ and $\epsilon$ to the 2D coordinates of each pose keypoint in $\mathcal{K}$. We then generate the pose maps as usual and train our models. The results of this experiment, conducted on the Smarthome CS protocol, are presented in Figure \ref{fig:noisy_poses}. As the reliability of the poses decreases, the accuracy of the model with PAAT quickly declines towards the baseline. On the other hand, while the accuracy of the model with PAAB also drops, it stabilizes at around $70\%$. These experiments show the crucial role of  poses in video understanding and demonstrate the robustness of PAAB and PAAT to considerable noise in pose information.

\begin{figure}[htbp]
    \vspace{-3pt}
    \begin{center}
    \scalebox{0.9}{
    \begin{minipage}{0.58\textwidth}
        \centering
        \captionof{table}{TimeSformer + our pose-aware methods, compared to the SOTA models on Toyota-Smarthome. Modality indicates the modalities required at inference time.}
        \label{tab:sota_smarthome}
        \resizebox{1\linewidth}{!}{
        \begin{tabular}{p{3cm} M{0.5cm} M{0.5cm} M{1cm} M{1cm} M{1cm} M{1cm}}
            \toprule
            \multirow{2}{*}{\textbf{Method}} & \multicolumn{2}{c}{\textbf{Modality}} & \multicolumn{3}{c}{\textbf{Accuracy (\%)}}\\
             & \textbf{RGB} & \textbf{Pose} & \textbf{CS} & \textbf{CV1} & \textbf{CV2} \\
            \toprule
            I3D \cite{i3d} & \ding{51} & \ding{55} & 53.4 & 34.9 & 45.1 \\
            VPN \cite{das2020vpn} & \ding{51} & 3D & 60.8 & 43.8 & 53.5 \\
            AssembleNet++ \cite{assemblenetplusplus} & \ding{51} & \ding{55} & 63.6 & - & - \\
            VPN++ \cite{vpn++} & \ding{51} & 3D & 71.0 & - & 58.1 \\
            MMNet \cite{mmnetTPAMI22} & \ding{51} & 2D & 70.1 & 37.4 & 46.6 \\
            \midrule
            Video Swin \cite{liu2021videoswin} & \ding{51} & \ding{55} & 69.8 & 36.6 & 48.6 \\
            MotionFormer \cite{motionformerNeurIPS21} & \ding{51} & \ding{55} & 65.8 & 45.2 & 51.0 \\
            
            \midrule     
            TimeSformer \cite{timesformer} & \ding{51} & \ding{55} & 68.4 & 50.0 & 60.6 \\
            \rowcolor{Gray}
            PAAB (ours) & \ding{51} & 2D & 71.4 & \textbf{54.9} & 61.8 \\
            \rowcolor{Gray}
            PAAT (ours) & \ding{51} & \ding{55} & \textbf{72.5} & 54.8 & \textbf{62.2} \\
            \bottomrule
        \end{tabular}
        }
        \label{}
    \end{minipage}
    \hfill
    \begin{minipage}{0.44\textwidth}
        \includegraphics[width=\linewidth]{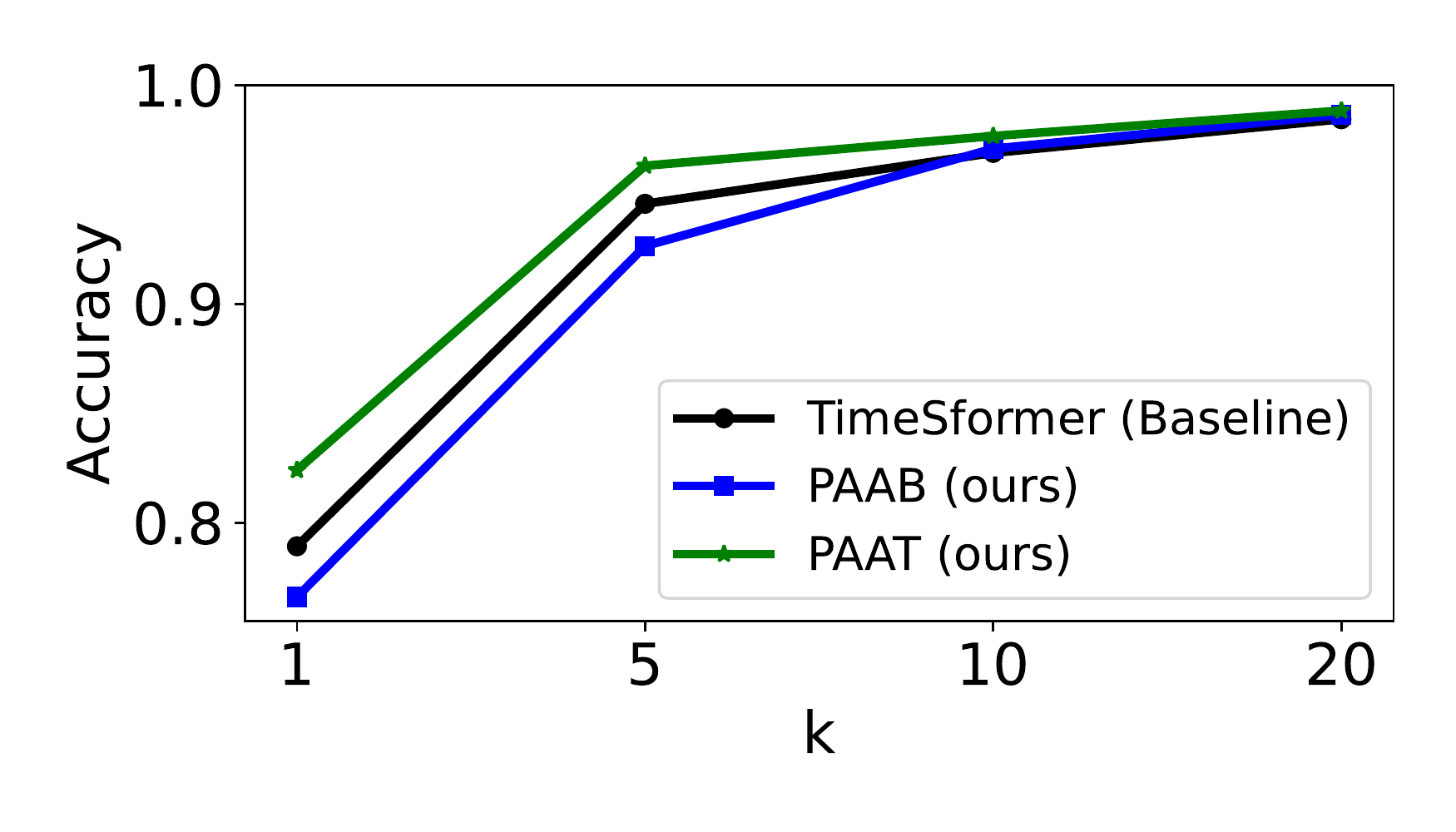}
        \vspace{-0.65cm}
        \caption{NTU (CS) pre-trained PAAT \& PAAB (with TimeSformer backbone) for $k$-NN video retrieval on NUCLA. Notably, PAAT performs better at $k=1,5,10$}
        \label{fig:ucla_video_retrieval}
    \end{minipage}}
    \end{center}
    \vspace{-10pt}
\end{figure}

\begin{table}[ht] \vspace{-0.15in}
    \centering
    \caption{State-of-the-art comparison on NTU-RGB+D and NUCLA. } 
    \begin{subtable}{0.48\linewidth}
        \centering
        \caption{TimeSformer + our pose-aware methods on NTU.}
        \resizebox{1\textwidth}{!}{
		\begin{tabular}{p{2.8cm} M{0.5cm} M{0.5cm} M{1.1cm} M{1.1cm} M{1.1cm} M{1.1cm}}
            \toprule
            \multirow{2}{*}{\textbf{Method}} & \multicolumn{2}{c}{\textbf{Modality}} & \multicolumn{3}{c}{\textbf{Accuracy (\%)}} \\
             & \textbf{RGB} & \textbf{Pose} & \textbf{CVS1} & \textbf{CVS2} & \textbf{CVS3} \\
            \toprule
            ST-GCN\cite{ghosh2018stacked_stgcn} & \ding{55} & 3D & 74.8 & 59.8 & 31.4 \\
            3D ResNet-50 \cite{can_spatio-temporal} & \ding{51} & \ding{55} & 83.9 & 67.9 & 42.9 \\
            S3D \cite{s3dECCV18} & \ding{51} & \ding{55} & 84.1 & 66.4 & 40.1 \\
            \midrule
            
            Video Swin \cite{liu2021videoswin} & \ding{51} & \ding{55} & \textbf{86.9} & 72.7 & 51.4 \\
            MotionFormer \cite{motionformerNeurIPS21} & \ding{51} & \ding{55} & 85.3 & 72.2 & 51.3 \\
            \midrule
            TimeSformer \cite{timesformer} & \ding{51} & \ding{55} & 83.5 & 72.6 & 50.3 \\
            \rowcolor{Gray}
            PAAB (ours) & \ding{51} & 2D & 85.8 & 73.1 & 51.6 \\
            \rowcolor{Gray}
            PAAT (ours) & \ding{51} & \ding{55} & 85.4 & \textbf{73.1} & \textbf{51.8} \\
            \bottomrule
        \end{tabular}
        }
    \end{subtable}%
    \hfill
    \begin{subtable}{0.49\linewidth}
        \caption{TimeSformer+our pose-aware methods on NUCLA.}
        \centering
        \resizebox{1\textwidth}{!}{
		\begin{tabular}{p{2.8cm} M{0.5cm} M{0.5cm} M{1cm} M{1cm} M{1cm} M{1cm}}
            \toprule
            \multirow{2}{*}{\textbf{Method}} & \multicolumn{2}{c}{\textbf{Modality}} & \multicolumn{3}{c}{\textbf{Accuracy (\%)}} \\
             & \textbf{RGB} & \textbf{Pose} & \textbf{CS} & \textbf{CV3} &\textbf{Avg}\\
            \toprule
            Glimpse Cloud~\cite{glimpse} & \ding{51} & \ding{55} & - & 90.1 & 87.6 \\
            VPN~\cite{das2020vpn} & \ding{51} & 3D & - & 93.5 & - \\
            VPN++~\cite{vpn++} & \ding{51} & 3D & - & 93.5 & - \\
            MMNet & \ding{51} & 3D & - & \textbf{93.7} & 88.7 \\
            \midrule
            Video Swin \cite{liu2021videoswin} & \ding{51} & \ding{55} & 90.7 & 89.6 & 84.3 \\
            MotionFormer \cite{motionformerNeurIPS21} & \ding{51} & \ding{55} & 90.2 & 89.4 & 88.4 \\
            
            \midrule
            TimeSformer \cite{timesformer} & \ding{51} & \ding{55} & 90.7 & 91.8 & 90.5 \\
            \rowcolor{Gray}
            PAAB (ours) & \ding{51} & 2D & 93.4 & 92.9 & \textbf{91.3} \\
            \rowcolor{Gray}
            PAAT (ours) & \ding{51} & \ding{55} & \textbf{95.4} & 92.7 & 90.8 \\
            \bottomrule
        \end{tabular}
        }
    \end{subtable}%
    \label{tab:sota_results} \vspace{-0.15in}
\end{table}

\subsection{Results}
In this section, we present the performance of our models across three downstream tasks, and present a state-of-the-art comparison with the representative baselines.

\textbf{Action recognition.}\quad We compare the performance of PAAB and PAAT to various state-of-the-art (SOTA) methods on the Smarthome, NTU, and NUCLA datasets in Table~\ref{tab:sota_smarthome} and Table~\ref{tab:sota_results} (a-b). 
Our results reveal that PAAB and PAAT set a new benchmark on the Smarthome dataset which presents the challenges of real-world scenarios, including challenges in pose estimation.  
For a fair comparison with our models, we have primarily included models that leverage the RGB modality for the NTU and NUCLA datasets. This is despite the popularity of pure-pose based methods in these datasets, as they are more representative to our evaluation scenario. The superior performance of our models, exhaustively evaluated on cross-view protocols, underlines the \textit{view- agnostic representation} learned by PAAB and PAAT through the use of 2D poses. 
Interestingly, despite solely relying on RGB, our models exhibit competitive results when compared with methods employing both RGB and 3D poses. This shows the capability of video transformers with PAAB or PAAT to capture \textit{pose-aware features}.
In addition, our models are compared with prominent video transformer models~\cite{liu2021videoswin, motionformerNeurIPS21, timesformer}. We find that either PAAB or PAAT, when employed with TimeSformer, surpasses the performance of state-of-the-art video transformers (except on CVS1 of NTU), boasting an absolute margin of up to 18.3\%.

\begin{figure}[htbp]
    \vspace{-3pt}
    \begin{center}
    \begin{minipage}{0.44\textwidth}
        \includegraphics[width=\textwidth]{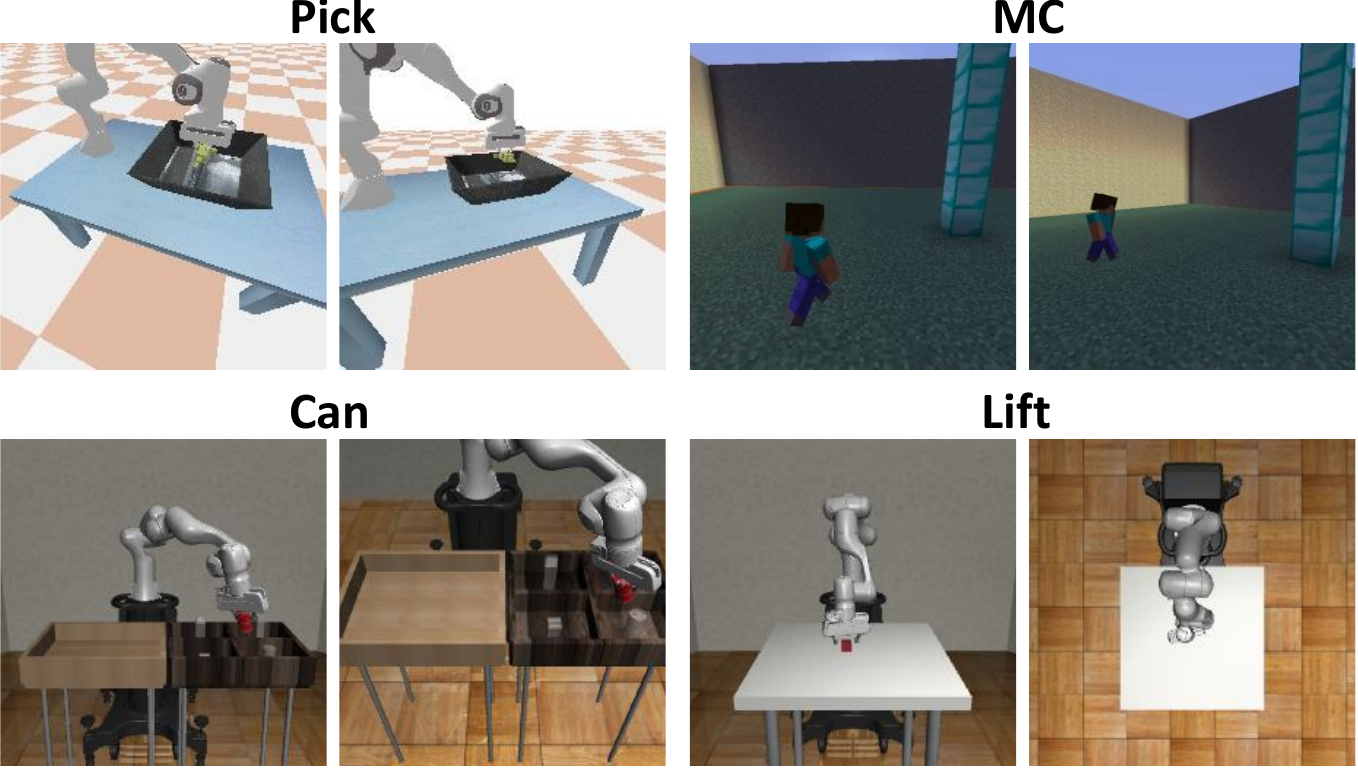}
        \captionof{figure}{Example video alignment frames from two different viewpoints.}
        \label{fig:alignment_datasets}
    \end{minipage}
    \hfill
    \begin{minipage}{0.52\textwidth}
        \centering
        \setlength{\tabcolsep}{5pt}
        \captionof{table}{DeiT + TCN + our pose-aware methods on multi-view robotic video alignment. Metric is alignment error.}
        \resizebox{\textwidth}{!}{
        \begin{tabular}{lc p{1cm} p{1cm} p{1cm} p{1cm} p{1cm}}
            \toprule
            \textbf{Method} & \textbf{Backbone}  & \textbf{Pick} & \textbf{MC} & \textbf{Can} & \textbf{Lift} \\
            \toprule
            TCN~\cite{sermanet2018time} & CNN   & 0.273 & 0.286 & - & - \\
            Disentangle~\cite{shang2021disentangle} & CNN &  0.155 & 0.233 & - & - \\ 
            
            3D TRL \cite{3dtrl} & ViT & \textbf{0.116} & 0.202 & 0.060 & 0.081 \\
            \midrule
            DeiT~\cite{deit}+TCN & ViT &   0.216 & 0.292 & 0.065 & 0.095 \\
            \rowcolor{Gray}
            PAAB (ours) & ViT & 0.165  & 0.151 & \textbf{0.053} & 0.075 \\
            \rowcolor{Gray}
            PAAT (ours) & ViT & 0.159 & \textbf{0.149} & 0.059 & \textbf{0.074} \\
            \bottomrule
        \end{tabular}
        }
        \label{tab:sota_video_align} 
    \end{minipage}
    \end{center}
    \vspace{-10pt}
\end{figure}

\textbf{Multi-view robotic video alignment.}\quad Table \ref{tab:sota_video_align} illustrates the performance of our models on the Pick, MC, Can, and Lift datasets. We present the alignment error (lower is better) for each method. Note that PAAB and PAAT are implemented in the DeiT encoder~\cite{deit} which is trained with TCN losses~\cite{sermanet2018time}.
We find that both PAAB and PAAT improves the baseline DeiT~\cite{deit} by 21.8\%  on the MC dataset, which is notable as it contains the largest viewpoint variation of all the datasets.  
While our models deliver superior results compared to most of the SOTA methods, they fall slightly short of 3DTRL~\cite{3dtrl} on the Pick dataset.

\textbf{Video retrieval.}\quad We demonstrate the generalizability of our models, by presenting the performance of our NTU pre-trained models for video retrieval on NUCLA, as illustrated in Figure~\ref{fig:ucla_video_retrieval}. However, PAAB's performance falls short in this context, while PAAT outperforms all. This shows the generalizability power of PAAT, attributed to its joint optimization strategy. 


\begin{figure}
    \centering
    \begin{subfigure}{0.48\textwidth}
        \includegraphics[width=\textwidth]{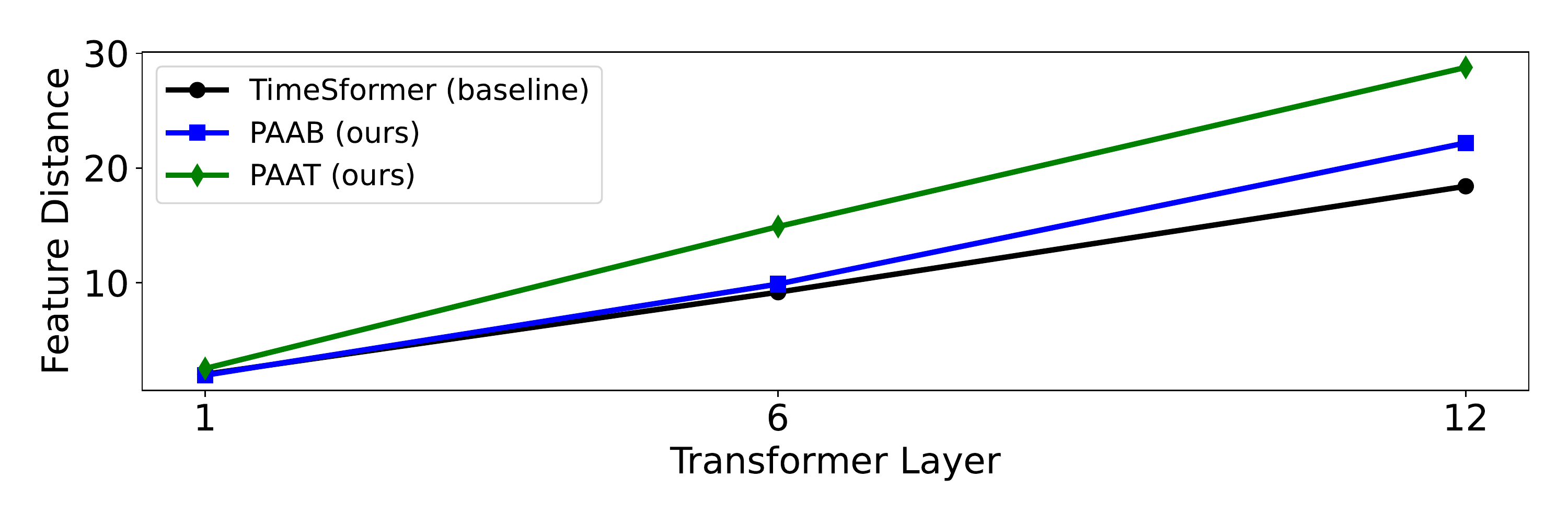}
        \caption{Average pose and non-pose token feature distance.}
        \label{fig:avg_dist_pose_bg}
    \end{subfigure}
    \begin{subfigure}{0.48\textwidth}
        \includegraphics[width=\textwidth]{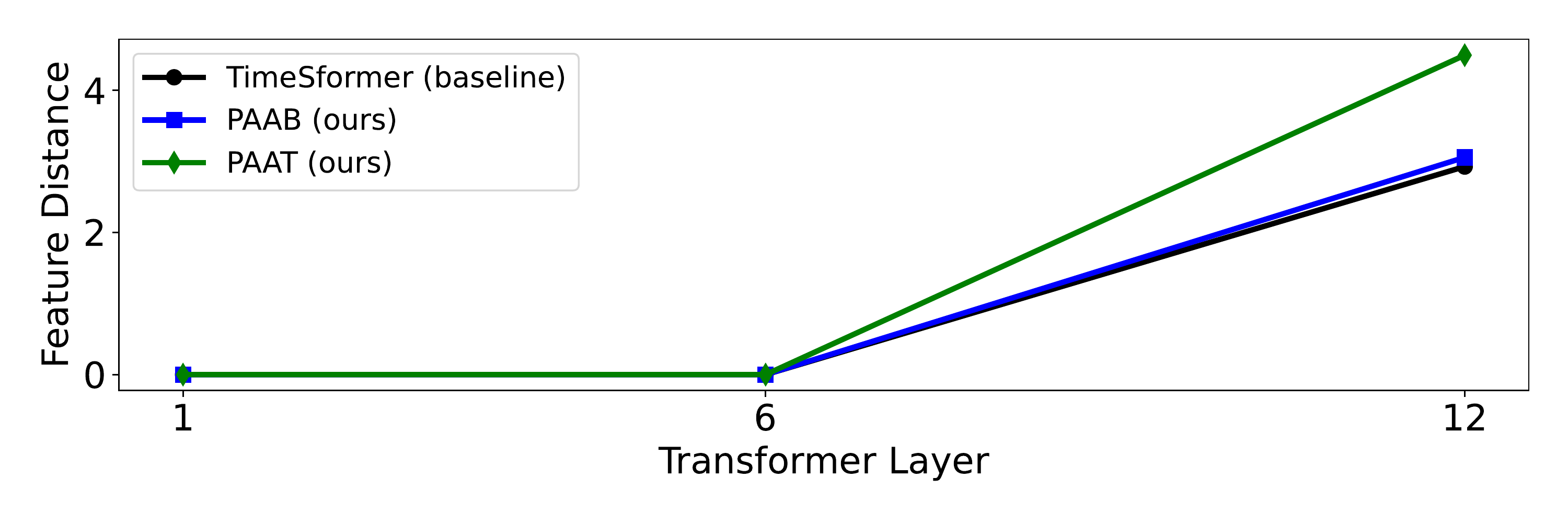}
        \caption{Average feature distance between pose tokens.}
        \label{fig:avg_dist_pose_pose}
    \end{subfigure}
    \hfill
    \begin{subfigure}{\textwidth}        
        \includegraphics[width=\textwidth]{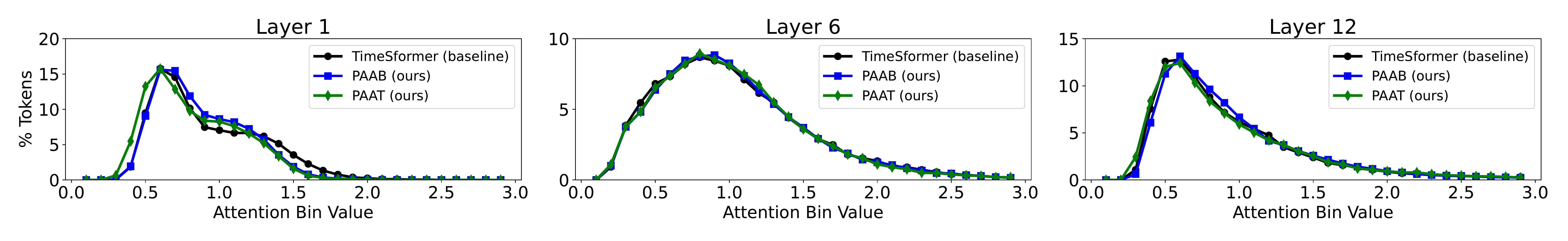}
        \caption{Attention distributions of layers 1, 6, and 12.}
        \label{fig:attention_distributions}
    \end{subfigure}
    \label{fig:model_diagnostics}  \vspace{-0.2in}
    \caption{\textbf{Feature \& Attention Analysis.} From (a) we see our methods learn to disentangle pose and non-pose tokens in the feature space. (b) shows that PAAT learns more separable pose tokens. (c) shows that interestingly, our methods maintain similar attention distributions to the baseline.} \vspace{-0.2in}
\end{figure}

\subsection{Feature \& Attention Analysis}\label{sec:model_diagnostics}
In this section, we explore the feature space and attention distributions learned by our models on a subset of the Smarthome. In Figure~\ref{fig:avg_dist_pose_bg}, we compute the average feature distance between the pose and non pose tokens in the feature space. Both PAAB and PAAT learn to better disentangle the pose and the non pose token representation compared to the baseline TimeSformer, with PAAT achieving superior feature separability due to its keypoint-specific prediction task. Figure~\ref{fig:avg_dist_pose_pose} further confirms this, where PAAT exhibits better separability of individual pose features, unlike PAAB. 
In addition to analyzing the feature space, we also explore the attention distributions learned by our methods. In Figure \ref{fig:attention_distributions} we report the percentage of tokens across various attention value bins at different layers. Interestingly, the attention distributions of our models align closely with the baseline, implying that our methods primarily leverage the feed-forward layers to learn pose-aware representations.




\section{Related Works}
\vspace{-0.05in}
In recent years, vision transformers~\cite{dosovitskiy2020vit, liu2021swin, deit, t2t, tnt, crossvit} have overtaken CNNs~\cite{resnet, vgg16, szegedy2016inception} in performance across numerous image-based tasks~\cite{dosovitskiy2020vit, carion2020detr, strudel2021segmenter}. Similarly, video transformers~\cite{timesformer, liu2021videoswin, vivit, motionformerNeurIPS21, mvit1, mvit2}  have had a comparable effect on 3DCNNs~\cite{x3d, lin2019tsm, i3d, C3D} and two-stream CNNs for video-based tasks~\cite{twostream, twostreamfusion, slow_fast}.
While these video transformers are tailored for analyzing web-based videos~\cite{kinetics, ucf, kuehne2011hmdb, AVA}, emphasizing prominent motion patterns and frame-centric actions, they often fall short when dealing with real-world videos. These videos~\cite{MSRDailyactivity3D, CAD-60, CAD-120, ntu120, NTU_RGB+D, smarthome, DML-smartactions, charades}, typically recorded in indoor settings and encompassing Activities of Daily Living (ADL), present challenges that these transformers are not designed to handle. The challenges of ADL typically includes subtle motion, videos captured from multiple camera viewpoints, and actions with similar appearance.
To address these challenges of ADL, studies~\cite{stgcn, msaagcn, hyunggun2022cvpr_infogcn, hachiuma2023unifiedskele} have been conducted on pure-pose based approaches that utilizes 2D and 3D poses. These approaches are effective on datasets recorded in laboratory settings~\cite{NTU_RGB+D, ntu120, nucla} where human actions are not spontaneous. However, they struggle with real-world videos~\cite{vpn++, smarthome} that necessitate appearance modeling of the scene to incorporate object encoding. In response, various approaches~\cite{smarthome, das2020vpn, vpn++, STA_hands, glimpse} have integrated RGB and pose modalities to model ADL. Notably, these methods typically utilize 3D poses, which are dependent on depth sensors or computationally intensive RGB algorithms~\cite{lcrnet_new, videopose3d}. 
In contrast, our methods, PAAB and PAAT, leverage 2D poses, which are generally accurate and easier to obtain.
The closest to our work~\cite{chainedcnn, pose_for_action} perform multi-tasking for pose estimation and action recognition by sharing a 3D CNN encoder with multiple heads. Unlike these, PAAT's multitasking (\textit{patch-keypoint prediction task}) is tailored for use in ViTs and intriguingly, it exhibits higher effectiveness at the initial layers rather than the final layers. To our knowledge, this is the first attempt to learn a pose-aware representation using vision transformers.



\vspace{-0.05in}
\section{Conclusion}
\vspace{-0.05in}


In conclusion, we proposed \textbf{PAAB} and \textbf{PAAT}, two methods for learning pose-aware representations with ViTs in the first attempt at combining the RGB and 2D pose modalities into a single-stream ViT. 
Based on our extensive experimental analysis, we find that incorporating pose information leads to generalized ViTs that are effective across multiple tasks, and even across datasets. Surprisingly, we observe that our methods do not significantly alter the attention distributions of the backbone ViTs, and instead rely on the feed-forward layers of the model to learn pose-aware representations. 

As for which method to use, \textit{we recommend end users to prefer PAAT over PAAB} due to its consistent superior performance, enhanced generalizability and its ability to learn more fine-grained pose representations. PAAT also demands less computational resources than PAAB during inference, necessitating no poses and additional computational parameters. However, PAAB can be a viable choice under conditions of poor pose quality and absence of computational constraints.


Future research will investigate the utilization of other entity-specific priors, like segmentation masks, to address a broad range of vision tasks.

\section*{Acknowledgments}
We thank the lab members of ML Lab at UNC Charlotte for valuable discussion. 
We thank Jinghuan Shang and Saarthak Kapse for their helpful feedback.
This work is supported by the National Science Foundation (IIS-2245652).

\bibliographystyle{plainnat}
\bibliography{refs}

\newpage
\section*{\Large Appendix}
\appendix

\section{Datasets and Protocol description}
\textbf{Action recognition}\quad For the task of action recognition we evaluate our methods on three popular Activities of Daily Living (ADL) datasets. \textbf{Toyota-Smarthome}\cite{smarthome} (Smarthome, SH) provides 16.1k video clips of elderly individuals performing actions in real-world settings. The dataset contains 18 subjects, 7 camera views, and 31 action classes. For evaluation, we follow the cross-subject (CS) and cross-view (CV1, CV2) protocols. Due to the unbalanced nature of the dataset, we use the mean class-accuracy (mCA) performance metric. The dataset provides 2D skeletons containing $13$ keypoints that were extracted using LCRNet \cite{lcrnet_new}, which we use to generate the pose maps for our method. \textbf{NTU-RGB+D}\cite{NTU_RGB+D} (NTU) provides 56.8k video clips of subjects performing actions in a laboratory setting. The dataset consists of 40 subjects, 3 camera views, and 60 action classes. For evaluation, we follow the cross-view-subject (CVS) protocols proposed in \cite{varol21_surreact} and evaluate performance with top-1 classification accuracy. In the CVS protocols, only the $0^\circ$ view from the CS training split is used for training, while testing is carried out on the $0^\circ$, $45^\circ$, and $90^\circ$ view from the CS test split, which are referred to as CVS1, CVS2, and CVS3. We use the CVS protocols because they better represent the cross-view challenge. The dataset provides 2D skeletons containing $25$ keypoints extracted using Randomized Decision Forest~\cite{37}, which we use to generate the pose maps. \textbf{Northwestern-UCLA Multiview activity 3D Dataset}\cite{nucla} (N-UCLA) contains 1200 video clips of subjects performing actions in a laboratory setting. The dataset consists of 10 subjects, 3 camera views, and 10 action classes. For evaluation, we follow the cross-view (CV) protocols in which the model is trained on two camera views and tested on the remaining view. For example, the CV3 protocol indicates the model was trained on views 1, 2 and tested on view 3. For evaluation, we report the accuracy on CV3 and the average accuracy on all cross-view protocols. We employ OpenPose \cite{OpenPose} to extract 2D skeletons containing $18$ keypoints and to generate the pose maps.

\textbf{Choice of Action Recognition Datasets and protocols}\quad Contrary to popular action recognition methods evaluated on datasets like Kinetics~\cite{kinetics} and SSV2~\cite{something}, our method targets scenarios emphasizing human poses, which we argue are crucial in Activities of Daily Living. This consideration influences our choice of datasets. Given that datasets like Kinetics often position humans centrally, close to the camera source, many poses remain obscured, thereby minimizing the relevance of skeletal data in these contexts~\cite{stgcn}.  

For evaluation on NTU dataset, we follow the CVS protocols since they are challenging owing to the disparity in the training distribution. While most methodologies evaluate this dataset using Cross-subject (CS) and Cross-view (CV) protocols, these tend to be less rigorous, saturated, and do not  reflect real-world scenarios. The performance of PAAB and PAAT enhance the baseline TimeSformer metrics on CS and CV protocols by a slight margin of 0.3\%-0.5\%, indicating that pose-aware RGB representations do not necessarily provide an additional performance boost. Nevertheless, the efficacy of the pose-aware representation, as learned from the pre-trained NTU (CS), is manifested in its generalizability for video retrieval tasks (see Fig. 5 in the main paper).

\section{Dataset specific Implementation details}
We train all of our video models (TimeSformer based) on $8$ RTX A5000 GPUs with a batch size of $32$ for Smarthome and $64$ for NTU and NUCLA. We train the image models (DeiT based) used in multi-view robotic video alignment on a single RTX A5000 GPU, with a batch size of $1$.

\textbf{Action recognition}\quad In all experiments, we follow a training pipeline similar to \cite{timesformer}. The RGB inputs to our models are video frames with a size of $8\times224\times224$ for Smarthome and NUCLA and a size of $16\times224\times224$ for NTU. Frames are sampled at a rate of $\frac{1}{32}$ for Smarthome and $\frac{1}{4}$ for NTU and NUCLA. To ensure that the video frames input to our model will contain pose keypoints, prior to sampling frames we first extract a $224 \times 224$ crop from the video that contains only the human subject. This can be done using the pose keypoints extracted from the RGB or by using a pre-trained human detector \cite{fasterrcnn_humandetector}. Our backbone model in which we insert PAAB and PAAT is a Kinetics-400 \cite{kinetics} pre-trained TimeSformer \cite{timesformer} model. For fine-tuning, we train the models for $15$ epochs.

\textbf{Multi-view robotic video alignment}\quad We use DeiT as a backbone architecture for inserting PAAT and PAAB and use the time-contrastive loss \cite{sermanet2017tcn_robotalign} to train our models. The training enforces that the distance between video frames are close if the frames are temporally close, but far if they are temporally distant. We train all of our models from scratch and follow the training recipe provided in~\cite{3dtrl}.

\section{Model Diagnosis}
\begin{figure}[htbp]
    \vspace{-3pt}
    \begin{minipage}{0.5\textwidth}
        \centering
        \caption{Effects of pre-training PAAB and PAAT with and without Kinetics 400 (K400) pre-training.}
		\begin{tabular}{c c c}
            \toprule
             \textbf{Method} & \textbf{Pretraining} & \textbf{mCA} \\
            \toprule
            PAAB & None & \textbf{71.42} \\
            PAAB & K400 & 69.98 \\
            \midrule
            PAAT & None & \textbf{72.54} \\
            PAAT & K400 & 72.25 \\
            \bottomrule
        \end{tabular}
		\label{tab:BOTH_pretraining_effects}
    \end{minipage}
    \hfill
    \begin{minipage}{0.46\textwidth}
        \centering
        \includegraphics[width=\textwidth]{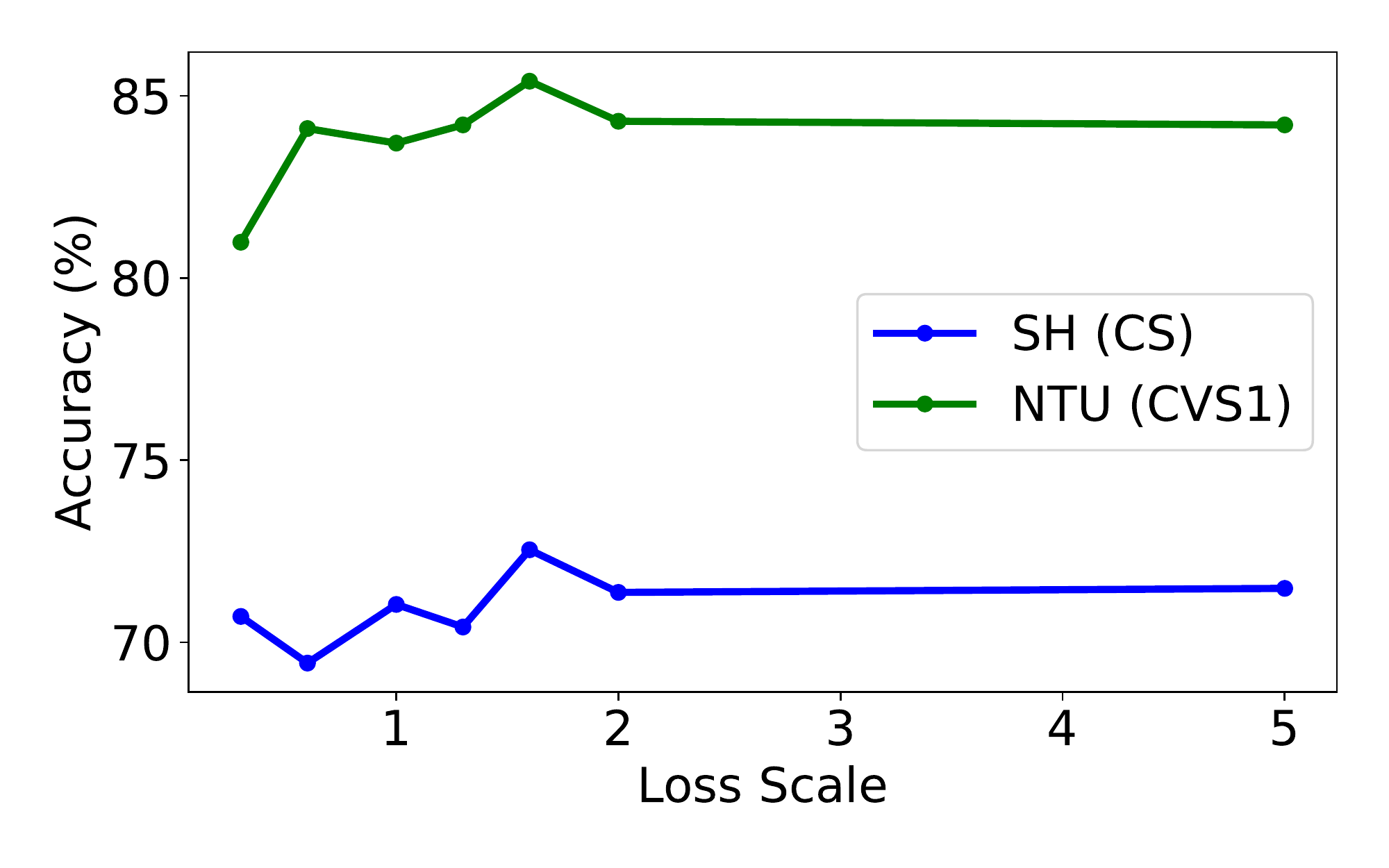}
        \caption{Ablation on the loss scale ($\lambda$) of PAAT}
        \label{fig:lossscale}
    \end{minipage}
    \vspace{-10pt}
\end{figure}

\textbf{Varying loss scale}\quad In Figure \ref{fig:lossscale} we present the results of varying PAAT's loss scaling factor, $\lambda$. We train PAAT on the Smarthome cross-subject (SH CS) and NTU CVS1 protocols with the following values of $\lambda$: $0.3, 0.6, 1.0, 1.3, 1.6, 2.0, 5.0$. We find that on both datasets, a value of $\lambda=1.6$ is optimal for training PAAT.

\textbf{Pre-training with Kinetics}\quad Surprisingly, our methods do not require Kinetics pre-training to achieve good performance. In Figure \ref{tab:BOTH_pretraining_effects}, we present the results of pre-training PAAB and PAAT on Kinetics-400~\cite{kinetics} prior to finetuning them on the Smarthome cross-subject protocol. Even more interesting is our observation that Kinetics pre-training degrades the performance of PAAB and PAAT. During Kinetics pre-training, we train the backbone without the incorporation of input poses. For PAAB, the additional block performs attention across all patches. The observed degradation in action classification performance may be attributed to discrepancies between pre-training and fine-tuning stages. This warrants the necessity of collecting more pose based real-world action recognition datasets.

\begin{figure}
    \centering
    \includegraphics[width=\textwidth]{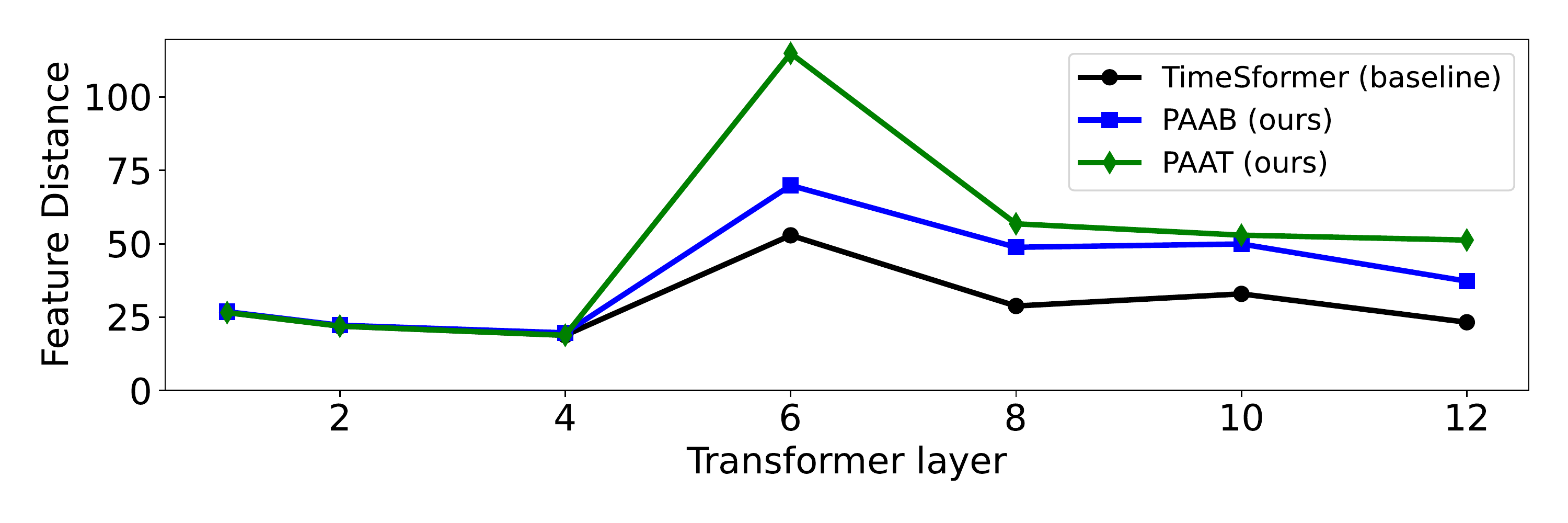}
    \caption{Average feature distance between tokens before and after the feed-forward networks.}
    \label{fig:ffn_analysis}
\end{figure}

\textbf{FFN feature distances}\quad In Figure~\ref{fig:ffn_analysis}, we show that PAAB and PAAT rely on the feed-forward networks (FFNs) within the model to learn pose-aware representations. We report the average feature distance between tokens before and after the FFNs at different layers of the backbone transformer. We find that in the initial layers, both PAAB and PAAT follow a similar trend to the baseline. However around layer $6$, we observe that the FFNs starts influencing the token representation. We recall that the attention distribution in PAAB and PAAT resembles with the baseline transformer. In this experiment, we find that both PAAB and PAAT alter the token representation more than the baseline, indicating that they are leveraging the FFNs rather than the attention distribution to learn pose-aware representations. This analysis reveals that the intermediate layers in transformers, particularly the Feed-Forward Networks (FFNs), play a pivotal role in learning pose-aware representation. However, this does not necessarily imply that PAAB and PAAT need to be integrated within these middle layers.

\section{Fusion of PAAB and PAAT}
While we have discussed PAAB and PAAT as two different strategies for learning pose-aware video representation, we also explore their combinination within a single architecture. In this experiment, we plug-in a PAAB following layer $12$ of the backbone TimeSformer and invoke a keypoint-classifier based PAAT block after layer $1$. Using a loss scaling factor of $\lambda = 1.6$, we observe a Smarthome (CS) action classification accuracy of 69.9\%, as compared to 71.4\% and 72.5\% achieved by PAAB and PAAT respectively. We argue that this drop in performance is owing to the conflicting gradients generated by the introduction of both modules (PAAB \& PAAT).

\section{Limitations} 
As previously mentioned, PAAB and PAAT are two distinct yet complementary methods for learning pose-aware representation, each possessing their unique strengths and weaknesses. The exploration of a strategy to integrate the benefits of both methods remains an open challenge. The amalgamation of both methods into a single model could potentially yield a more robust framework compared to individual models. 

Additionally, another key limitation of PAAB and PAAT is their dependency on pose data during training. Although recent advances have facilitated pose extraction from RGB~\cite{OpenPose}, the associated computational costs remain a challenge for training our methodologies. Moreover, PAAB necessitates pose information during inference, along with the inclusion of additional parameters, thereby extending the inference time.
\end{document}